\documentclass{article}
\usepackage{spconf,amsmath,graphicx,hyperref}
\usepackage{comment}
\usepackage{orcidlink}
\usepackage[utf8]{inputenc}
\usepackage{textcomp}

\title{Multi modal perceived conversation success prediction for spontaneous autistic and non-autistic dyadic interactions}
\name{Thanushi Withanage\thanks{This work is supported by Grand Challenge grant from University of Maryland}$^{\dagger\,\orcidlink{0000-0002-1869-1124}}$, Desi Jones$^{\star\,\orcidlink{0000-0001-6779-2477}}$, Noah Sasson$^{\ddagger\,\orcidlink{0000-0002-3676-1253}}$, Elizabeth Redcay$^{\star\,\orcidlink{0000-0002-1568-3102}}$, Carol Espy-Wilson$^{\dagger\,\orcidlink{0000-0002-1012-183X}}$}
  
\address{$^{\dagger}$Department of Electrical and Computer Engineering, University of Maryland College Park, USA  \\
      $^{\star}$Department of Psychology, University of Maryland College Park, USA \\
      $^{\ddagger}$Department of Psychology, University of Texas at Dallas, USA }

\let\OLDthebibliography\thebibliography
\renewcommand\thebibliography[1]{
  \OLDthebibliography{#1}
  \setlength{\parskip}{0pt}
  \setlength{\itemsep}{0pt plus 0.3ex}
}

\begin{document}
%
\maketitle

\begin{abstract}
\vspace{-0.2cm}
Conversation success is important for communication, social bonds, and well-being and is often reduced in autistic-neurotypical conversations. Previous studies have predicted perceived conversation success using acoustic and lexical features in pediatric autism-enriched samples, or using nonverbal behavior in non-autistic group interactions. Other work has characterized prosodic, facial, motor, and turn-taking synchrony in autistic interactions, but has primarily targeted diagnostic classification or group differences. The present study instead predicts participant perceived conversation success from complementary articulatory, facial, acoustic-prosodic, and turn-taking measures across three spontaneous dyadic-interaction datasets including autistic and non autistic individuals. By leveraging multimodal features, we demonstrate that conversational success is inherently multidimensional and requires a multi-faceted analytical framework. Combining these complementary modalities offers the potential to improve prediction of perceived conversation success by capturing information unavailable from any single modality.
\end{abstract}
\vspace{-0.3cm}
\vspace{-0.5cm}
\section{Introduction}
\label{sec:intro}
\vspace{-0.3cm}

Successful conversations requires partners to coordinate speech, facial expressions, and speaking turns while exchanging information and establishing social connection. However, perceived conversation success encompasses several dimensions, including enjoyment, interpersonal liking, and shared understanding, which may not be captured by a single behavioral measure \cite{pagel}. Previous studies have measured perceived conversation quality at both individual and group levels, linking higher ratings to balanced participation, fewer interruptions, and synchronized body language \cite{raman}. These findings motivate a multimodal approach to examining complementary verbal and nonverbal behaviors to understand how partners across different neurotypes experience their interactions.

Acoustic-prosodic entrainment, through which conversational partners adapt their vocal behavior, provides one potential marker of interaction quality. Prior research distinguishes proximity, convergence, and synchrony and associates acoustic adaptation with perceived social behavior \cite{levitan}. Pitch and intensity synchrony have also been associated with communicative success in task-oriented interactions involving typical, accented, and dysarthric speech \cite{borrie}. Extending this work to spontaneous conversations, our previous study identified associations between perceived success and acoustic, facial, and turn-taking measures \cite{withanage}. Nevertheless, identifying behavioral associations does not establish their ability to predict success in unseen conversations.

Computational studies have begun addressing this prediction problem. Müller et al. combined facial, prosodic, and interactional features to detect low rapport in small-group discussions, finding facial behavior particularly informative \cite{muller}. Yang et al. predicted conversation quality from acoustic and lexical features in an autism-enriched pediatric sample, with pretrained audio representations improving classification performance \cite{yang}. These studies demonstrate predictive potential but leave questions about spontaneous interactions across different neurotypes and generalization across datasets.

Autistic and non-autistic interactions provide an informative context because behavioral coordination may relate differently to perceived success across partner compositions. Studies report differences in prosodic entrainment and conversational timing \cite{wehrle}, while speech and turn-taking features have supported diagnostic classification \cite{plank}. However, neurotype-matching research shows that autistic people can experience strong rapport with autistic partners \cite{crompton}. Moreover, synchrony vs. rapport correlations vary across dyad types, and observers’ evaluations can differ from participants’ experiences \cite{efthimiou,jones}. These findings motivate predicting perceived success directly rather than inferring it from diagnostic differences or assuming that greater synchrony universally indicates better interaction.

Articulatory coordination offers an additional perspective on conversational adaptation. Correlation-based eigen spectra have been used to characterize coordination among speech-production subsystems in autism \cite{talkar}. Building on this representation, changes in articulatory coordination during conversation may provide information beyond conventional prosodic and visual measures in predicting conversational success.

In this study, we investigate multimodal prediction of perceived conversation success across three datasets of spontaneous dyadic interactions involving autistic and non-autistic participants. We integrate articulatory coordination, facial action units, pitch, intensity, and turn-taking features. Our objectives are to assess the predictive contributions of individual feature families and their combination, examine variation across interaction populations, and evaluate the consistency of behavioral predictors across datasets.

\vspace{-0.5cm}
\section{Datasets}
\vspace{-0.3cm}
We examined spontaneous dyadic conversations from three datasets spanning different age groups, neurotype compositions, interaction durations, and recording settings.
\vspace{-0.4cm}
\subsection{CANDOR dataset}
\vspace{-0.3cm}
The CANDOR Corpus (Conversation: A Naturalistic Dataset of Online Recordings) \cite{reece} was developed by BetterUp Labs and researchers at the University of Pennsylvania in 2023. It contains more than 1,650 unstructured, approximately 30-minute Zoom conversations between previously unacquainted adults in which we are choosing 500 conversations for this study. Participants ranged from 19 to 66 years of age and represented diverse demographic, ethnic, and educational backgrounds. Separate audio recordings were available for each speaker, eliminating the need for speaker diarization. Our analysis included sessions identified by the dataset's quality annotations as free from substantial background noise and interruptions.
\vspace{-0.4cm}
\subsection{UTD dataset}
\vspace{-0.3cm}
We analyzed a consented UTD subset \cite{jones,jones_intersectional} of 62 dyads: 18 autistic--autistic (AT--AT), 31 autistic--non-autistic (AT--NA), and 13 non-autistic--non-autistic (NA--NA). Autistic participants met diagnostic criteria on the Autism Diagnostic Observation Schedule, Second Edition. Participants were recruited through the University of Texas at Dallas and the surrounding community. Previously unacquainted partners, uninformed of each other's diagnostic status, engaged in a five-minute, face-to-face, unstructured get-to-know-you conversation.
\vspace{-0.4cm}
\subsection{UMD dataset}
\vspace{-0.3cm}
The UMD dataset comprises 12 spontaneous dyadic interactions involving autistic and non-autistic adults recruited from the University of Maryland student community who are previous acquainted and talking about familiar topics. Conversations were conducted through Zoom and lasted approximately 15 minutes.

\vspace{-0.5cm}
\section{Method}
\vspace{-0.3cm}
\subsection{Feature extraction}
\vspace{-0.2cm}
\subsubsection{Pitch and intensity}
\vspace{-0.2cm}
Audio was converted to mono and resampled to 16~kHz to standardize processing across datasets. For CANDOR dataset speaker-specific recordings were used and for the other two datasets speaker assignments and temporal boundaries were obtained from transcripts or diarization using pyannote \cite{pyanote}. Speech was segmented into turns using the corresponding timing annotations.

Fundamental frequency ($F_0$) trajectories were extracted from each turn using the Pitch-Estimating Neural Network (PENN) \cite{morrison}. Pitch values were normalized to reduce between-speaker differences. Speech intensity was extracted using Praat. For each turn, the \textbf{minimum, maximum, and mean values of pitch and intensity} were calculated.

Acoustic proximity was summarized using feature differences between adjacent turns produced by different speakers (\textbf{FDA}) and between non-adjacent turns produced by different speakers (\textbf{FDNA}) in the same conversation \cite{levitan, withanage}. These differences were averaged separately for each acoustic statistic over the full conversation and its two halves which characterizes local acoustic proximity relative to a non-adjacent-turn reference.

\vspace{-0.4cm}
\subsubsection{Facial action-unit synchrony}
\vspace{-0.2cm}
Facial movements were quantified using OpenFace \cite{baltruvsaitis}, which estimates frame-level facial action-unit (FAU) intensities from video. We used the 17 intensity-based FAUs available in the default output \cite{withanage}.

Synchrony was estimated separately for each FAU using windowed, lagged Pearson correlations between the two speakers' FAU trajectories. At a frame rate of 30 fps, the trajectories were divided into non-overlapping 5-second windows. Within each window, correlations were evaluated across temporal lags from $-5$ to $+5$ seconds in 0.5-second increments to accommodate delayed facial responses. The maximum correlation across the evaluated lags was retained as the window-level synchrony estimate. These coefficients were Fisher-$z$ transformed and averaged across valid windows to obtain one \textbf{conversation-level synchrony score per FAU}. Higher values indicate stronger positive temporal correspondence between partners' facial movements.

\vspace{-0.4cm}
\subsubsection{Turn-taking}
\vspace{-0.2cm}
Turn-taking behavior was characterized using the annotated start and end times of both speakers’ speech turns. Individual turn durations were calculated as the difference between each turn’s onset and offset, while the analyzed conversation duration spanned the beginning to the end of the whole interaction duration, including speech and pauses. Inter-turn silence was calculated as a positive gap between one speaker’s turn offset and the other speaker’s subsequent turn onset; overlapping transitions were not counted as silence. The resulting features were \textbf{total number of turns; minimum, maximum, and mean turn duration; and total and mean inter-turn silence}. 

Total conversation duration (\textbf{Duration}) was analyzed separately from the turn-taking modality because it reflects the overall length of the interaction rather than how speakers exchange turns. Although participants were instructed to converse for a specified period, actual durations varied slightly around the allotted time. These deviations may partly reflect participants’ willingness to continue talking. Analyzing duration separately allowed us to assess its predictive contribution independently of turn-taking behavior.
\vspace{-0.4cm}
\subsubsection{Articulatory coordination}
\vspace{-0.2cm}
A speech inversion system \cite{tabatabaee} was applied to each audio segment to estimate 10 time-varying channels: six vocal tract variables describing lip aperture, lip protrusion, tongue-tip constriction degree and location, and tongue-body constriction degree and location, together with nasalance, periodicity, aperiodicity, and pitch. Articulatory coordination feature (ACF) matrices were constructed from auto- and cross-correlations among these channels over multiple temporal delays \cite{seneviratne}. The eigenvalues of each ACF matrix were sorted in descending order to form a 150-dimensional eigenspectrum representing the distribution of variance across coordination dimensions.

Following our previous articulatory coordination framework \cite{thanushi}, each conversation was divided into two temporal halves. Segment-level eigenspectra were averaged separately for each speaker within each half, yielding 4 mean eigenspectra per conversation. Each mean eigenspectrum was summarized using the geometrically weighted score($\alpha$) to obtain weighted sum of geometric decay score (\textbf{WSGD}):
\vspace{-0.5cm}
\begin{equation}
WSGD_{s}^{(h)}
=
\sum_{i=1}^{D}
\bar{\lambda}_{s,i}^{(h)}\alpha^{i-1},
\qquad s,h\in\{1,2\},
\label{eq:wsed}
\end{equation} where $\bar{\lambda}_{s,i}^{(h)}$ denotes the $i$th eigenvalue of speaker $s$'s mean eigenspectrum in half $h$, and $D$ (150) is the spectrum dimensionality. The parameter $\alpha$ controls the decay of the weights, emphasizing the largest eigenvalues when $\alpha<1$. Our earlier study used $\alpha=0.51$ \cite{withanage}; the same formulation extends here.

Scores were rescaled to $[0,1]$, and the normalized values for both speakers in both halves were retained as four predictors. These features capture each speaker's coordination profile and its evolution during the interaction.

\vspace{-0.4cm}
\subsection{Perceived conversation success}
\vspace{-0.2cm}
Perceived conversation success (PCS) was derived from participants' post-conversation ratings of their interaction and conversational partner. Because the questionnaires differed across datasets, principal component analysis (PCA) was performed separately for each dataset to identify the dimensions underlying these evaluations and select the questions that best represented perceived interaction quality covering relevant aspects of conversational experience, including affect, enjoyment, interpersonal liking, and common ground. Before PCA, responses were standardized within each question to account for differences in rating scales. Questions with factor-loading magnitudes greater than $0.4$ were retained for PCS computation \cite{withanage}. Thus, question selection was specific to each dataset while targeting the broader construct of perceived conversation success.

Retained questions were oriented so that higher values indicated more favorable evaluations. Their standardized responses were averaged to obtain a participant-level composite, which was subsequently rescaled to $[0,1]$. Conversation-level PCS was calculated by averaging the two participants' scores.  This procedure accommodates differences in questionnaire content while providing a common numerical range for prediction.

\vspace{-0.4cm}
\subsection{Multimodal prediction system}
\vspace{-0.2cm}

PCS was modeled as a continuous outcome using ridge regression. Features from five modalities; WSGD, FAU synchrony, pitch, intensity, and turn-taking were concatenated into a single feature vector \(\mathbf{x}_i\) for each conversation. These features were standardized (centered and scaled using training-set means and standard deviations) and weighted by \(1/\sqrt{p_m}\), where \(p_m\) is the number of features in modality \(m\), to account for differences in modality dimensionality. Ridge regression estimated the intercept \(b\) and feature coefficients \(\boldsymbol{\beta}\) by minimizing the following objective:
\vspace{-0.4cm}
\[
\underset{b,\boldsymbol{\beta}}{\operatorname{minimize}}
\;\underbrace{\sum_{i=1}^{n}
\left(y_i-b-\mathbf{x}_i^\top\boldsymbol{\beta}\right)^2}_{\text{prediction error}}
+
\underbrace{\alpha\sum_{j=1}^{p}\beta_j^2}_{L_2\text{ penalty}}.
\]
\vspace{-0.4cm}

where $y_i$ is the actual PCS and
$b+\mathbf{x}_i^\top\boldsymbol{\beta}$ is the predicted PCS.
The model minimizes the sum of squared prediction errors plus
an $L_2$ penalty (the sum of squared regression coefficients),
balancing prediction accuracy with coefficient shrinkage.
Larger $\alpha$ encourages stronger shrinkage toward zero.

Performance within CANDOR was evaluated using five-fold nested cross-validation repeated five times. Each outer split used 400 of the 500 conversations for training and 100 for testing. Four-fold inner cross-validation divided the 400 training conversations into 300 for fitting and 100 for validation, selecting \(\alpha\) from nine logarithmically spaced values between \(10^{-4}\) and \(10^{4}\) by minimizing average validation mean absolute error (MAE; the average absolute difference between predicted and actual PCS). Preprocessing was fitted separately within each training fold to prevent information leakage. After selecting \(\alpha\), the model was refitted on the full outer training partition and evaluated on the held-out fold.

For external evaluation, the penalty strength was selected using cross-validation within CANDOR, and a final model was fitted to all eligible CANDOR conversations. This model was then applied to UTD and UMD using the same features, CANDOR-derived standardization, modality weights, and fitted coefficients, without retraining or tuning on the external datasets. 

MAE was the primary evaluation metric, supplemented by root mean squared error, \(R^2\), Pearson’s correlation, and Spearman’s correlation. CANDOR metrics were calculated from out-of-fold predictions within each repetition and summarized across repetitions; external metrics were calculated separately for each dataset. The multimodal model was compared with single-modality models, leave-one-modality-out ablations, and a training-mean baseline, using identical outer partitions for CANDOR comparisons.

\vspace{-0.5cm}
\section{Results}
\vspace{-0.1cm}

\begin{table}[t]
\centering
\caption{MAE of modalities for three datasets. Bold indicates the best performing model with lowest MAE. The baseline predicts the CANDOR training mean.}
\label{tab:mae}
\small
\begin{tabular}{lccc}
\hline
Model & CANDOR & UTD & UMD \\
\hline
Training-mean baseline & 0.1143 & 0.1471 & 0.2391 \\
Turn-taking & 0.1104 & \textbf{0.1183} & 0.2427 \\
Pitch & $\approx$0.1145--0.1147 & 0.1451 & 0.2408 \\
Intensity & $\approx$0.1145--0.1147 & 0.1612 & 0.2423 \\
WSGD & $\approx$0.1145--0.1147 & 0.1553 & 0.2403 \\
FAU synchrony & $\approx$0.1145--0.1147 & 0.1498 & 0.2387 \\
Duration only & 0.1103 & 0.1197 & 0.2375 \\
Five modalities, no duration & 0.1130 & 0.1602 & 0.2413 \\
Five modalities + duration & \textbf{0.1089} & 0.1267 & \textbf{0.2368} \\
\hline
\end{tabular}
\end{table}
\vspace{-0.3cm}
\subsection{Within-dataset prediction}
\vspace{-0.3cm}
On CANDOR, the full model achieved an MAE of $0.1089\pm0.0011$, compared with $0.1143\pm0.0009$ for the training-mean baseline, a relative reduction of 4.7\% (Table~\ref{tab:mae}). Variability denotes standard deviation across repetitions. Turn-taking was the strongest individual behavioral modality (MAE = 0.1104), closely matching duration alone (0.1103). The other individual modalities did not outperform the baseline. Combining the five behavioral modalities without duration yielded a smaller improvement (MAE = 0.1130).

The full model explained a modest proportion of PCS variance ($R^2=0.0957$; Table~\ref{tab:metrics}). Removing duration produced the largest increase in MAE, whereas removing turn-taking slightly improved performance (Table~\ref{tab:ablation}). Thus, turn-taking was informative alone but provided limited additional benefit when the remaining predictors and duration were included.

\begin{table}[t]
\centering
\caption{Additional performance metrics for the best performing model (lowest MAE).}
\label{tab:metrics}
\small
\begin{tabular}{lccc}
\hline
Metric & CANDOR & UTD & UMD \\
\hline
RMSE & 0.1378 & 0.1601 & 0.2671 \\
$R^2$ & 0.0957 & $0.0370$ & 0.0190 \\
Pearson $r$ & 0.3147 & $0.2350$ & 0.2858 \\
Spearman $\rho$ & 0.3019 & $0.2916$ & 0.2098 \\
\hline
\end{tabular}
\end{table}
\vspace{-0.5cm}
\subsection{Cross-dataset prediction}
\vspace{-0.3cm}

Cross-dataset evaluation showed that models trained exclusively on CANDOR achieved lower MAE than the CANDOR training-mean baseline on both UTD and UMD, with the strongest improvements observed on UTD (Table~\ref{tab:mae}). These results provide initial evidence that selected conversational features retain predictive value across differences in recording settings, interaction durations, and participant populations.

On UTD, the full multimodal model reduced MAE by 13.9\% relative to the baseline. Turn-taking, identified as the strongest individual behavioral modality through CANDOR cross-validation, achieved an even larger reduction of 19.6\%, yielding the lowest MAE among the primary models. Its positive $R^2$, Pearson correlation, and Spearman correlation further suggest that conversational timing captured transferable information about differences in perceived success. Duration alone also performed well (MAE = 0.1197), highlighting the value of temporal interaction characteristics. Also the full model improved absolute-score accuracy relative to the source-training-mean baseline, capturing conversation-level variation in UTD. To further interpret the strong performance of the turn-taking model on UTD, we examined turn-taking patterns at the extremes of the PCS distribution. In the same neurotype dyads we observed less number of turns, high mean turn duration and less inter-turn silence in high PCS conversations compared to low PCS conversations with high statistical significance. These descriptive findings suggest that conversational timing is highly relevant to perceived success, with patterns differing across dyad types.

On UMD, the full model achieved the lowest MAE among the primary models (0.2368), improving on the baseline by approximately 1.0\% and outperforming both duration alone (0.2375) and turn-taking alone (0.2427). The full model also yielded positive Pearson and Spearman correlations ($r=0.2858$, $\rho=0.2098$). Turn-taking showed particularly promising associations with observed PCS ($r=0.5860$, $\rho=0.5455$), indicating potential for ordering conversations by perceived success even though its absolute-score error exceeded the baseline. Exploratory ablation analysis further identified a lower MAE of 0.2303 when pitch was excluded, corresponding to a 3.7\% reduction relative to the baseline. Because this configuration was identified retrospectively on the external test set, it motivates independent validation of feature selection for cross-dataset transfer.

\begin{table}[t]
\centering
\caption{Change in MAE after removing each feature group from the full model: $\Delta\mathrm{MAE}=\mathrm{MAE}_{\mathrm{ablated}}-\mathrm{MAE}_{\mathrm{full}}$. Positive values indicate worse performance after removal.}
\label{tab:ablation}
\small
\begin{tabular}{lccc}
\hline
Removed group & CANDOR & UTD & UMD \\
\hline
Duration & $+0.00406$ & $+0.03353$ & $+0.00447$ \\
Turn-taking & $-0.00020$ & $+0.00097$ & $+0.00620$ \\
Pitch & $+0.00009$ & $+0.00705$ & $-0.00645$ \\
Intensity & $+0.00056$ & $-0.00404$ & $+0.00030$ \\
WSED & $+0.00055$ & $-0.00309$ & $-0.00182$ \\
FAU synchrony & $+0.00066$ & $-0.00410$ & $+0.00181$ \\
\hline
\end{tabular}
\end{table}
\vspace{-0.5cm}
\section{Conclusion}
\vspace{-0.4cm}
These findings highlight conversational timing as a promising transferable source of information and show that the full multimodal model achieved baseline-relative MAE improvements in both external datasets. The differing contributions of individual modalities suggest opportunities to improve generalization through feature selection and calibration across interaction contexts. In future we hope extend this analysis to multi speaker interactions among different neurotypes. 

\bibliographystyle{IEEEbib}
\bibliography{strings,refs}

\end{document}